\documentclass[10pt,letterpaper]{article}
\usepackage[top=0.85in,left=1.3in,footskip=0.65in,marginparwidth=1.3in]{geometry}

\usepackage[utf8]{inputenc}

\usepackage{cite}
\usepackage{soul,color}
\usepackage{nameref,hyperref}

\usepackage[right]{lineno}

\usepackage{microtype}
\DisableLigatures[f]{encoding = *, family = * }

\raggedright
\usepackage{changepage}

\usepackage[aboveskip=1pt,labelfont=bf,labelsep=period,singlelinecheck=off]{caption}

\makeatletter
\renewcommand{\@biblabel}[1]{\quad#1.}
\makeatother

\usepackage{tabularx}
    \newcolumntype{L}{>{\raggedright\arraybackslash}X}

\usepackage{lastpage,fancyhdr,graphicx}
\usepackage{epstopdf}
\usepackage{float}
\usepackage{natbib}
\usepackage{xcolor}
\usepackage{amsmath}
\usepackage{filecontents}
\fancyheadoffset[L]{2.25in}
\fancyfootoffset[L]{2.25in}

\usepackage{color}

\definecolor{Gray}{gray}{.25}

\usepackage{mathtools, nccmath}
\usepackage{mathrsfs}
\usepackage{mathptmx}
\usepackage{wrapfig}
\usepackage[pscoord]{eso-pic}
\usepackage[fulladjust]{marginnote}
    
\usepackage{url}
\usepackage{xcolor}
\definecolor{newcolor}{rgb}{.8,.349,.1}
\usepackage{mathtools, nccmath}
\usepackage{multicol, blindtext}
\usepackage{graphicx}
\usepackage{mathrsfs}
\usepackage{rotating}
\usepackage{blindtext}
\usepackage{mathptmx}
\usepackage{anyfontsize}
\usepackage{t1enc}
\usepackage{amsmath, amssymb, txfonts}
\usepackage{amsmath}
\usepackage{rotating}
\usepackage{lipsum}
\usepackage[document]{ragged2e}
\usepackage{caption,graphicx,newfloat}
\DeclareCaptionType{InfoBox}
\usepackage{placeins}
\usepackage{longtable, lscape}
\usepackage{hyperref}
\hypersetup{
    colorlinks=blue,
    linkcolor=blue,
    filecolor=magenta,      
    urlcolor=cyan,
}

\usepackage{amssymb}
\usepackage{latexsym}
\usepackage{mathtools}
\usepackage{url}
\usepackage{xcolor}
\usepackage{graphicx}
\usepackage{multicol}    
\usepackage{color}    
    
\begin{document}

\begin{flushleft}
{\Large
\textbf\newline{UGRWO-Sampling for COVID-19 dataset: A modified random walk under-sampling approach based on graphs to imbalanced data classification}
}
\\

\bigskip

Saeideh Roshanfekr\textsuperscript{1},
Shahriar Esmaeili\textsuperscript{2}, 
Hassan Ataeian\textsuperscript{3}, and
Ali Amiri\textsuperscript{3}
\\
\bigskip
\bf{1} {Department of Computer Engineering and Information Technology, Amirkabir University of Technology, 424 Hafez Avenue, 15875-4413 Tehran, Iran}
\\
\bf{2} {Department of Physics and Astronomy, Texas A$\&$M University, 4242 TAMU, University Dr., College Station, TX 77840, US}
\\
\bf{3} {Department of Computer Engineering, University of Zanjan, University Blvd., 45371-38791 Zanjan, Iran}\\

\bigskip

\end{flushleft}
\begin{abstract}

This paper proposes a new RWO-Sampling (Random Walk Over-Sampling) based on graphs for imbalanced datasets. In this method, two schemes based on under-sampling and over-sampling methods are introduced to keep the proximity information robust to noises and outliers. After constructing the first graph on minority class, RWO-Sampling will be implemented on selected samples, and the rest will remain unchanged. The second graph is constructed for the majority class, and the samples in a low-density area (outliers) are removed. Finally, in the proposed method, samples of the majority class in a high-density area are selected, and the rest are eliminated.Furthermore, utilizing RWO-sampling, the boundary of minority class is increased though the outliers are not raised. This method is tested, and the number of evaluation measures is compared to previous methods on nine continuous attribute datasets with different over-sampling rates and one data set for the diagnosis of COVID-19 disease. The experimental results indicated the high efficiency and flexibility of the proposed method for the classification of imbalanced data.

\end{abstract}

\newpage

\section{Introduction} \label{sec1}
Classification is one of the most critical issues in data mining, pattern recognition, and machine learning. Numerous classification methods have been successful in various applications but, when these methods are performed on the imbalanced data sets, the performance of minority class may not be very satisfactory \citep{ataeian_2019}. If the distribution of samples in these two classes is unequal, then data sets are considered imbalanced. In this paper, the class with more samples is named 'majority class,' and the other class is called 'minority class.'
The imbalanced data sets have various applications such as network intrusion detection \citep{Giacinto_2008, roshanfekr_2019}, credit scoring \citep{Schebesch_2008}, spam filtering \citep{Tang2008SpamSD}, text categorization \citep{Zheng_2004} and anomaly detection\citep{Pichara_2011}. Data sets are divided into data with all continuous attributes, all-discrete attributes, and continuous and discrete attributes (hybrid attributes). There are two solutions in order to deal with these data sets. First, setting the distribution of data. Second, setting the classifier to operate with these data sets. Numerous techniques have been proposed for this solution. The most common method is sampling (i.e., the data sets are balanced by reducing or increasing the size of data sets). These two methods are called "under-sampling" and "over-sampling," respectively. The under-sampling method is straightforward, but it eliminates useful samples of the majority class, whereas the over-sampling method increases the risk of over-fitting \citep{Batista_2004}. Over-sampling is the opposite of the under-sampling method. It duplicates or interpolates minority samples in the hope of reducing the imbalance. The over-sampling method assumes the neighborhood of a positive sample to be still positive, and the samples between two positive samples positive \citep{Japkowicz_2002, Laurikkala_2002, Ling_1998, Kotsiantis_2006, Yen_2009, Sun_2009}.

The common sampling methods are RO-Sampling (Random Over-Sampling) and RU-Sampling (Random Under-Sampling) \cite{barandela2004imbalanced}. RO-Sampling balances the distribution of data by duplicating the samples of the minority class randomly. However, RU-Sampling can eliminate some useful samples of the majority class randomly. SMOTE (Synthetic Minority Over-Sampling Technique) method is the over-sampling method that generates new synthetic samples along the line between the minority samples and their nearest neighbors. MSMOTE (Modified SMOTE) method is a modified SMOTE method by classifying the sample of minority class into three groups, which are security samples, border samples, and latent noise samples, and this method runs different strategies for any groups \citep{Hu_2009}. Tomek links \citep{Tomek_1976} refers to a method for identifying pairs of nearest neighbors in a dataset that have different classes. Removing one or both of the examples in these pairs (such as the examples in the majority class) makes the decision boundary in the training dataset less noisy or ambiguous. Tomek links can be used as an under-sampling method or data cleaning method. As an under-sampling method, only samples belonging to the majority class are eliminated, and as a data cleaning method, samples of both classes are removed.

Even though over-sampling minority class samples can balance class distributions, other issues normally found in data sets with skewed class distributions aren't solved. Often, class clusters are not well described when you consider that some majority class samples might be invading the minority class space. The opposite also can be genuine because interpolating minority class samples can enlarge the minority class clusters, introducing synthetic minority class samples too deeply in the majority class space. Inducing a classifier under such a situation can lead to over-fitting.  To create better-defined class clusters, in paper  \citep{Batista_2004}    proposed applying Tomek links to the over-sampled training set as a data cleaning method. Thus, samples from both classes are removed instead of removing only the majority of class samples that form Tomek links.

\cite{Yen_2009} presented a cluster-based under-sampling approach. First, all the samples are divided into some clusters in this approach. The main issue is that there are different clusters in a dataset, and each cluster appears to have unique characteristics. For example, suppose a cluster has more minority class samples than several majority class samples. In that case, it does not have the characteristics of the majority class samples and behaves more like the minority class samples. Therefore, this method selects a suitable number of majority class samples from each cluster by considering the ratio of the number of majority class samples to the number of minority class samples in the cluster. \cite{Zhang_2014} proposed the RWO-Sampling method by generating some synthetic minority class samples via randomly walking from the data. The synthetic minority class expands the minority class boundaries since this method does not change the data distribution.

\cite{Han_2005} presented borderline-SMOTE1 and borderline-SMOTE2. These methods are based on SMOTE. The borderline samples of the minority class are more easily misclassified
than those far from the borderline. Thus, these methods only over-sample the borderline samples of the minority class. \cite{Barua_2014} presented a Majority Weighted Minority Oversampling Technique (MWMOTE) approach. The technique uses the samples of the majority class near the decision boundary to select the samples of the minority class effectively. It then assigns the weight to the samples chosen according to their importance in learning. The samples closer to the decision boundary are given more weights than others. 

AdaBoost increases the weights of misclassified instances and decreases those correctly classified using the same proportion, without considering the imbalance of the data sets \citep{Freund_1996}. Therefore, traditional boosting algorithms do not function well in the minority class.  An improved boosting algorithm is proposed by \cite{Joshi}, which updated the weights of positive prediction differently from the weights of negative predictions. When dealing with imbalanced data sets, the class boundary learned by Support Vector Machines (SVMs) is apt to skew toward the minority class, thus increasing the misclassified rate of the minority class. A class boundary alignment algorithm is also proposed by \cite{Wu_2003}, which modifies the class boundary by changing the kernel function of SVMs.

\cite{Bunkhumpornpat_2011} presented Density-Based Minority Over-sampling Technique DBSMOTE). They generated synthetic instances along the shortest path from each positive instance to a minority-class cluster with a graph. \cite{Perez_Ortiz_2015} proposed three methods are based on analyzing the data from a graph-based perspective in order to easily include the ordering information in the synthetic pattern generation process. \cite{Luo_2014} presented a k-nearest neighbor (KNN) weighting strategy for handling the problem of class imbalance. They proposed CCW (class confidence weights) that uses the probability of attribute values given class labels to weight prototypes in KNN. In this work, we only discuss over-sampling and under-sampling approaches, and we try to increase the efficiency of the RWO method using hybrid methods.
 
This study presents a modified RWO-Sampling \citep{Zhang_2014} by constructing two local graphs. At first, the samples of the minority class in high-density regions are selected by constructing a proximity graph (proximity graph with k-nearest neighbors), then the RWO-Sampling method is implemented. New samples from the minority class are generated without being affected by noises and outliers owing to these being either out of the graph or on the boundary of the graph. The main difference between noises and outliers is that noise is any undesirable or unwanted signal or part of a signal, and noise may or may not be random, but an outlier is a data point or value that differs considerably from most other data in a dataset. In the second graph in UGRWO sampling, the majority of class samples in high-density areas are selected, and the rest of the samples are eliminated.  The samples of the high majority class, which are noises and outliers, are mostly eliminated. We implement four classifiers to compare the performance of the proposed method with the RWO-Sampling, SMOTE \citep{Chawla_2011}, MWMOTE \cite{Barua_2014}, and RO-Sampling \citep{Batista_2004}. The performance of these classifiers was evaluated in terms of common metrics, such as F-measure, G-mean, accuracy, AUC, and TP rate. The experiment has been performed on nine benchmark UCI data sets with different skew degrees and one COVID-19 blood test data set.

This article is organized into five sections. First, section \ref{sec2} explains the RWO-Sampling method and proposes the method on how to handle the class imbalance problem. Then, the results of the experiments on nine real datasets from UCI  and one COVID-19 dataset and performance estimators of the proposed approach are discussed in Section \ref{sec3}. Finally, Section \ref{sec4} includes the conclusion of this research work.

\section{Proposed method}\label{sec2}
\subsection{Background}
In the RWO-Sampling method, consider the training dataset $T$, and the minority class instance set $P= \{x_{1},..., x_{n}\}$. Each $x_{j}$  represented by m attributes is an m-dimensional vector representing a point in the m-dimensional space. The attribute set is named $A= \{a_{1},..., a_{m}\}$, and $a_{i}(j)$ is used to denote the value of attribute $a_{i}$ for instance $x_{j}$. The RWO-Sampling method acts for continuous datasets and discrete datasets in different ways. For discrete attributes, this method uses roulette wheels to generate synthetic values for them and continuous attributes; the method is shown in Algorithm \ref{tab1} is used \citep{Zhang_2014}.

According to the central limit theorem, this method generates synthetic minority class samples for unknown data distribution problems. For a multiple attribute dataset, the mean and standard variance for each attribute using the minority class data denote $\mu_{i}$ and $\sigma_{i}$ for ith attribute $a_{i}$. Each attribute can be considered a random value, and each attribute value can be considered its one sampling value. $\mu_{i}'$ and $\sigma_{i}'$ denote its real mean and standard deviation for random variable $a_{i}$. In this case, if the number of the minority class samples methods is infinite, then it gives us the following term;

\begin{equation}\label{eq1}
    \frac{\mu_{i}-\mu_{i}'}{\sigma_{i}'/\sqrt{n}}\rightarrow N(0, 1).
\end{equation}
 If Eq. \ref{eq1} is satisfied, we will have the following equation
 \begin{equation}\label{eq2}
     \mu_{i}'=\mu_{i}-r\times \frac{\sigma_{i}'}{\sqrt{n}}
 \end{equation}
 where r is a sampling value of distribution N(0, 1).
 
 \begin{table}[thb] 
    \centering
    \begin{tabular}{|l|}
        \hline
 input: T, M (M is over sampling rate)\\
output: $M\times n$  synthetic instances for minority class\\
for $i=1$ to m\\
\quad if $a_{i}$ is a continuous attribute\\
\qquad calculating the mean $\mu_{i}=\frac{\sum_{j=1}^n a_{i}(j)}{n}$\\
\qquad and the variance $\sigma_{i}^2 = \frac{1}{n} \sum_{j=1}^n (a_{i}(j) - \mu_{i})^2$\\
\quad if $a_{i}$ is a discrete attribute
\qquad calculating the occurrence probability for each value of $a_{i}$\\
while $(M>0)$\\
\quad for each $x_{j} \in  p$\\
\qquad for each $a_{i} \in  p$\\
\quad \qquad if $a_{i}$ is a continuous attribute\\
\qquad \qquad generating a random value $a'_{i}(j)=a_{i}(j) - \frac{\sigma_{i}}{\sqrt{n}} N(0, 1)$\\
\qquad \quad if $a_{i}$ a hybrid attribute\\
\qquad \qquad generating a random value for attribute $a_{i}$ using roulette\\
\qquad forming a synthetic instance $(a'_{1}(j), a'_{2}(j),…, a'_{m}(j))$\\
\qquad \qquad \qquad \qquad \qquad \qquad M=M-1\\
return the $M\times n$ instances for the minority class\\\hline
    \end{tabular}\caption{Algorithm 1- RWO-Sampling (T, M: a positive integer) \citep{Zhang_2014}.}\label{tab1}
\end{table}
 
This method keeps the data distribution unchanged and balances different class samples by creating synthetic samples by randomly walking from the real data. When some conditions are satisfied, it can be proved that both the expected average and the standard deviation of the generated samples are equal to that of the original minority class data \cite{Zhang_2014}. 

\subsection{UGRWO-Sampling approach}
\cite{Zhang_2014} proposed the RWO-Sampling method that expands the minority class boundary after synthetic samples are generated. However, it does not generate new samples around the mean point of the real samples of the minority class since it increases the likelihood of over-fitting. In this paper, we improved the disadvantages of the RWO method, which increases the over-fitting and does not generate synthetic samples around the mean point. This method eliminates the impact of samples that probably are noises and outliers, so synthetic samples are generated around the mean point. We presented a modified version of the Random Walk Over-sampling method from the RWO method. The proposed method is attempted to find a solution to the problem of RWO-Samplings. Therefore, utilizing the KNN method, the UGRWO-Sampling creates synthetic samples around the mean points since by eliminating the impact of noises or outliers, the new samples are generated around the mean, which decreases the probability of over-fitting. This method is based on two graphs. These two graphs are introduced to keep the proximity information. An independent graph is constructed for each class, which is called the local majority graph and minority local graph.

\begin{figure*}[ht]
	\centerline{\includegraphics[width=1.1\textwidth,clip=]{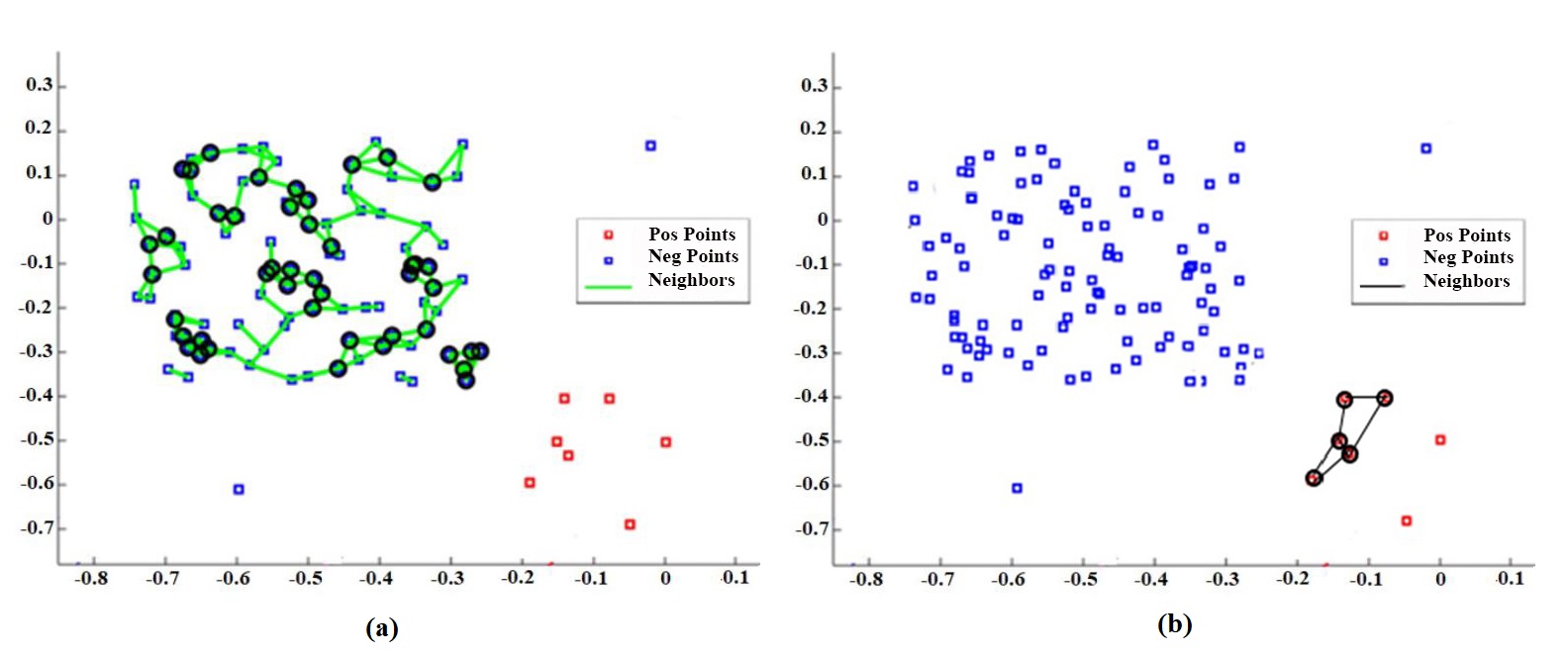}}
	\caption{Representation of the decreasing the number of majority (a) and minority (b) classes.}\label{fig1}
\end{figure*}

In the first step, we construct the local majority graph. If the corresponding two vertices are K-Nearest Neighbors (KNN) of each other, then an edge is added between a pair of vertices (Fig. \ref{fig1}). The graph's vertices with K or more degrees will be retained, and the rest will be deleted. The second graph is made in the second step, and the edges are created as before. Again, the graph's vertices with K or more degrees will be retained, and RWO-sampling runs on them.

Specifically, the adjacent matrix of the majority class, which is indicated by $U$ which is defined as follows,

\begin{equation}\label{eq3}
U_{ij} = \left\{
\begin{array}{rl}
\tau_{ij}, & \qquad x_{i} \in N_{k}(j)\quad and \quad x_{j} \in N_{k}(i) \\
0 & \qquad otherwise.\\
\end{array} \right.
\end{equation}
where $(N_{k}(j))$ is a set of the k-nearest neighbors in the majority class of the point $x_{j}$, $(N_{k}(j))$ k-nearest neighbors of the point $x_{i}$, and $U$ is adjacent matrix. $\tau_{ij}$ is a a scalar value or any characters for showing k-neareast neighbors in a special vertex. $\tau_{ij}$ can be assumed any amount expect zero, and $i,j=1,..,n$. Then we define the under-sampling coefficient as \citep{roshanfekr_2019},

\begin{equation}\label{eq4}
u_{i} = \left\{
\begin{array}{rl}
1, & \qquad \sum_{j} U_{ij}\geq k \\
0 & \qquad otherwise\\
\end{array} \right.
\end{equation}

The sample $x_{i}$ with non-zero $u_{i}$ can be selected, and the rest of the samples are deleted. In this way, samples that may be noise or outliers will be eliminated. Generally, samples in high-density regions have more chance of becoming non-zero degree vertices, while samples in low-density regions (e.g., outliers) increase the likelihood of isolated vertices with zero degrees. The method shown in Algorithm \ref{tab2} is used.

\begin{table}[thb]
	\caption{Algorithm 2- UGRWO-Sampling (T,M,k: a positive integer)}\label{tab2}
	\centering
	\begin{tabular}{|l|l|}
		\hline
		& Input: T, M (k is parameter of KNN) \\ 
		& Output: remaining samples of majority class\\ 
		& \qquad Constructing majority local graph\\
		&\qquad \qquad Defining adjacent matrix, $U_{ij}$, for samples of each class\\
		&\qquad \qquad \quad If ($x_{i} \in N_{k}(j)$\quad and \quad $x_{j} \in N_{k}(i)$)\\
		\textbf{step 1}    & \qquad \qquad \quad $U_{ij} = \tau_{ij}$\quad else\quad $U_{ij} = 0$ \\
		&\qquad \qquad Defining the under-sampling coefficient, $u_{i}$,for samples of each class\\
		&\qquad \qquad \quad If $\sum_{j} U_{ij}$ is greater than or equal to k\\
		&\qquad \qquad \quad $u_{i}=1$ \quad else \quad $u_{i}=0$\\
		&\qquad \qquad Deleting samples of majority class with zero $u_{i}$\\
		&\qquad Return samples of majority class with nonzero  $u_{i}$\\\hline 
		&   Input: T, M, k (M is over sampling rate and k is parameter of KNN)\\ 
		&Output: the synthetic samples for the minority class\\
		&\qquad Constructing minority local graph\\
		&\qquad \qquad Defining adjacent matrix, $U_{ij}$, for samples of each class\\
		&\qquad \qquad \quad If ($x_{i} \in N_{k}(j)$\quad and \quad $x_{j} \in N_{k}(i)$)\\
		&\qquad \qquad \quad $U_{ij} = \tau_{ij}$\quad else\quad $U_{ij} = 0$\\
		&\qquad \qquad Defining the under-sampling coefficient, $u_{i}$,for samples of each class\\
		&\qquad \qquad \quad If $\sum_{j} U_{ij}$ is greater than or equal to k\\
		&\qquad \qquad \quad $u_{i}=1$ \quad else \quad $u_{i}=0$\\
		\textbf{step 2}&\qquad Selecting samples of minority class with nonzero $u_{i}$ and using the following steps for them\\
		&\qquad \qquad \quad for i=1 to m\\
		&\qquad \qquad \qquad Calculating the mean $\mu_{i}=\frac{\sum_{j=1}^n a_{i}(j)}{n}$\\
		&\qquad \qquad \qquad  and the variance $\sigma_{i}^2 = \frac{1}{n} \sum_{j=1}^n (a_{i}(j) - \mu_{i})^2$\\
		&\qquad \qquad \quad while (M>0)\\
		&\qquad \qquad \qquad for each $x_{j} \in p$\\
		&\qquad \qquad \qquad \quad for each $a_{j} \in A$\\
		&\qquad \qquad \qquad \qquad  Generating a random value $a'_{i}(j)=a_{i}(j) - \frac{\sigma_{i}}{\sqrt{n}} N(0, 1)$\\
		&\qquad \qquad \qquad Forming a synthetic instance $(a'_{1}(j), a'_{2}(j),…, a'_{m}(j))$\\
		&\qquad \qquad \qquad \qquad \qquad \qquad  \qquad \qquad M= M-1\\
		&Return the synthetic samples for the minority class\\\hline
	\end{tabular}
\end{table} 

The local minority graph is constructed as follows: At first, the samples of minority class with non-zero $u_{i}$ are selected, and then the RWO-Sampling method is run on the selected samples (see Figure \ref{fig1}). Then, the two local graphs eliminate the impact of noise and outliers in the RWO-Sampling method. In this way, a new sample will not be generated from the samples that may be noises or outliers.

\begin{figure*}[ht]
	\centerline{\includegraphics[width=1.1\textwidth,clip=]{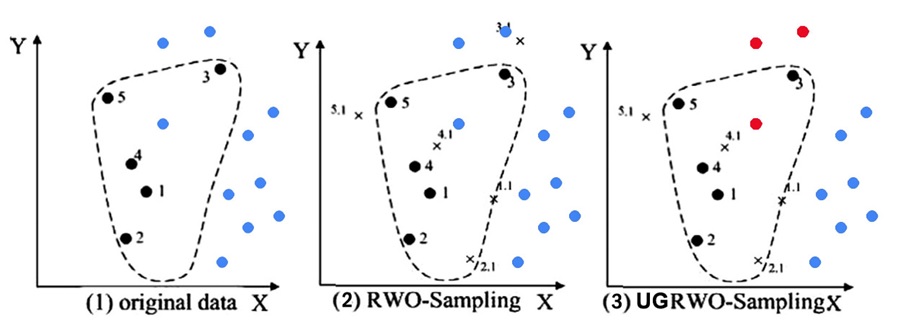}}
	\caption{Representing of different over-sampling approaches. Black dots represent the minority samples, blue and red points denote majority samples and the other points represent the synthetic data due to the over-sampling. (1) Shows the original data. (2) RWO-Sampling, (3) UGRWO-Sampling.}\label{fig2}
\end{figure*}

In Figure \ref{fig2}, the dashed closed line around the original data represents the boundary of the minority class data. RWO-Sampling uses a random walk to generate synthetic samples. Thus, it can expand the positive class border and increase the positive class classification accuracy. Our method separates the positive and negative samples, and the black points of the minority class samples and the red and blue points are the majority sample classes. In the algorithm, sample 3 is considered in the majority class as noise or outlier, and therefore the RWO algorithm will not run on it, and sample 3.1 will not be generated. In the majority class, red points are considered noises and eventually eliminated.

In subsection 3. 2, the proposed method is compared with the GRWO-Sampling method. In comparison to the RWO-Sampling, the advantage of our method is that the outlier sample will not be increased.

\section{Experimental results}\label{sec3}
The proposed approach is applied to the synthetic dataset with different conditions and nine benchmark datasets from the UCI Repository\footnote{http://www.ics.uci. edu/~ mlearn/MLRepository. html} and one COVID-19 dataset \footnote{https://www.kaggle.com/einsteindata4u/covid19} in Section \ref{ssec3_1}. All the classifiers are implemented in MATLAB 7.0 environment on a PC with an Intel P4 processor with 4 GB RAM. Section 3.2 summarizes the performance obtained for all the proposed approaches.
\subsection{Description of datasets and evaluation metrics}\label{ssec3_1}
 \subsubsection{Synthetic Dataset}
We first evaluate the UGRWO-sampling method on synthetic binary class data sets. Then, three groups of synthetic data sets are generated. 

	1) Overlap: The between-class distance and IR are adjusted.

	2) Noise: The noise level and IR are adjusted.

	3) Disjunct: The number of small disjuncts and IR are adjusted. Small disjuncts are those disjuncts that classify a few training samples. These disjuncts are interesting because they have a much higher error rate than large disjuncts.

\begin{figure}
	\centering
	\includegraphics[width=1.1\linewidth]{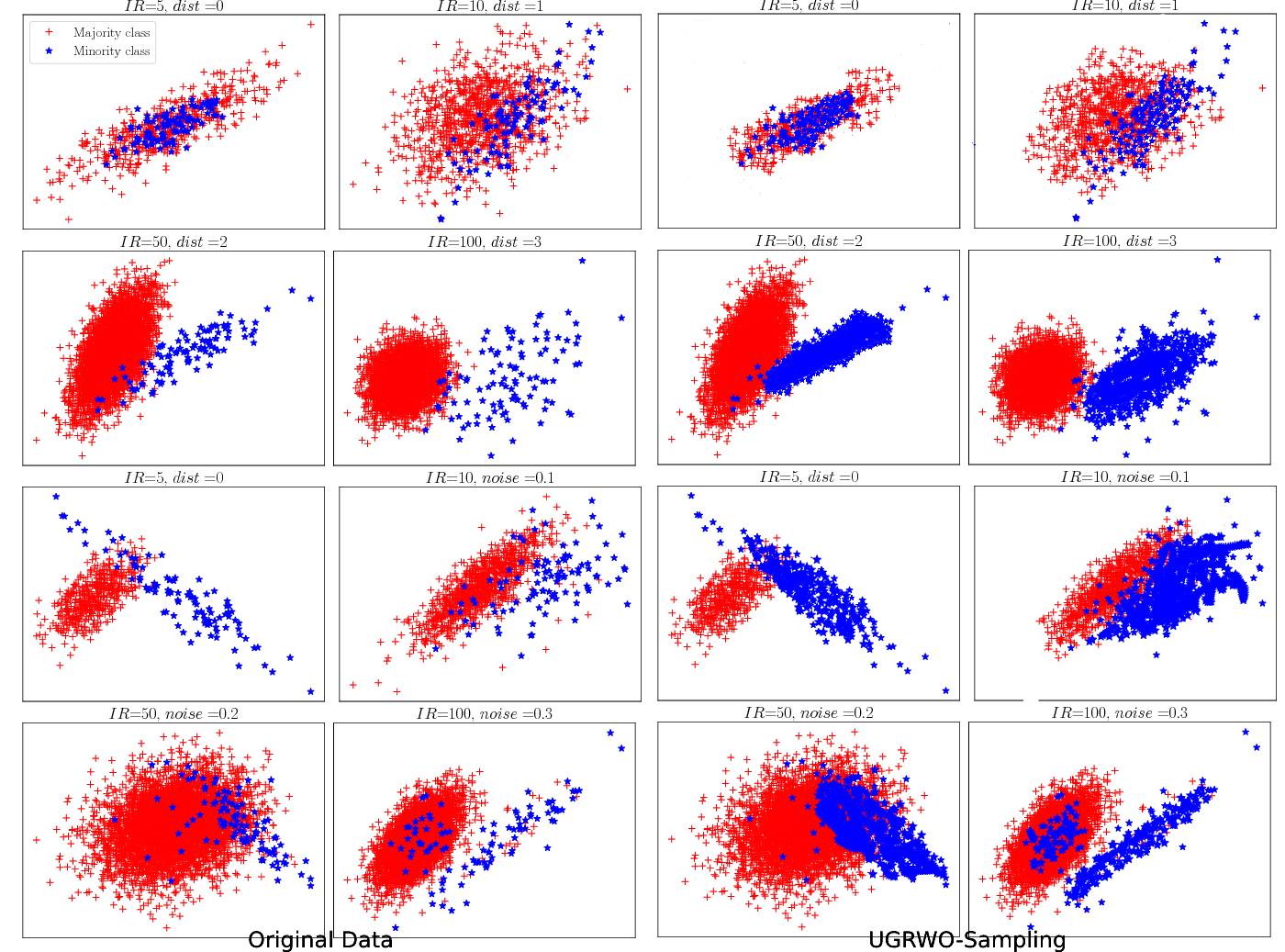}
	\caption{synthetic dataset and effect UGRWO-Sampling on it}
	\label{fig4}
\end{figure}

All data sets have two classes generated from a normal distribution with two dimensions. The number of samples in the minority class is fixed at 100, and the number of samples in the majority class varies in the set $\left\lbrace 500, 1000, 5000\right\rbrace $, where IRs are 5, 10, and 50, respectively. For data set group overlap, the distance between two classes dist varies in the set $\left\lbrace 0,1,2,3 \right\rbrace$ and there is no noise. For data set group Noise, the noise level noise varies in the set $\left\lbrace 0, 0.1, 0.2, 0.3\right\rbrace$, where 0.1 means that $10\%$ of the minority class samples are labeled as the majority class and that the same number of the majority class samples are labeled as the minority class. The distance between the two classes for the data set group Noise is fixed at two. For data set group Disjunct, the number of small disjuncts of each class disjunct varies in the set $\left\lbrace 1, 2, 4, 8\right\rbrace $. For example, disjunct $=$ 2 means each class has two disjuncts. The distance between adjacent disjuncts is set at two. The position of the majority class and the minority class with the different numbers of disjuncts is shown in Figure \ref{fig4}. For all synthetic datasets, the covariance matrix for each class is set to
\begin{center}
	\begin{equation*}
	\sigma= \left[ {\begin{array}{cc}
			a_{11} & a_{12} \\
			a_{21} & a_{22} \\
	\end{array} } \right] + 0.1I
\end{equation*}
\end{center}

where a11, a22 $=$ $\left\lbrace 0,1\right\rbrace $ and a12, a21 $=$ $\left\lbrace -1,1 \right\rbrace $ are uniformly random numbers. The extra term 0.1I ensures that the covariance matrix is positive semidefinite. The covariance matrix for the positive and negative classes is set differently, and the covariance matrix is drawn ten times to produce different combinations.
Therefore, there are two groups	of 120 datasets with different degrees of overlapping, different IR, different noise level, and different covariance. Four of the datasets in dataset group Overlap and four of the datasets in dataset group Noise and the result of performing UGRWO on the datasets are shown in Figure \ref{fig4}.

According to the Figure \ref{fig4}, it can be seen that by implementing a new method on the database with different rates of imbalance and conditions such as noise and overlap, the new method has been able to eliminate the majority class noise and increase the minority class samples in the noise-free range.
\subsubsection{Real Benchmark Dataset}
The UGRWO-sampling and other methods are applied to ten benchmark datasets, and these benchmark datasets represent a wide range of fields, number of instances, and positive class labels. Table \ref{tab3} gives the characteristics of these datasets, the minority class for each dataset is shown in Table \ref{tab3}, and the rest of the classes are majority class.

In SMOTE, SMOTE$+$Tomek, MWMOTE, GRWO-Sampling, and UGRWO-Sampling, the K-parameter in the KNN method is selected from the set  $\left\lbrace 3,5,10,15 \right\rbrace$. The over-sampling rate is 100, 200, 300, 400, and 500\% for data with continuous attributes. If an attribute is a missing value, the mean of that attribute is considered for it. A 10-fold cross-validation scheme is used to evaluate the performance of each over-sampling method. For the above procedure, we should implement the selected baseline algorithms under different over-sampling rates by toolboxes of MATLAB. 

The performance of a classifier is evaluated based on metrics such as accuracy \citep{Provost_1998}, Geometric Mean (G-mean) \citep{Kubat_1997}, F-measure \cite{Wu_2005}, AUC, and TPrate. In this paper, TP and FP are the numbers of true positive and false positive, respectively, and TN and FN are the numbers of a true negative and false negative, respectively. For the minority class data, its precision$=TP/(TP+FP)$ and TPrate=$TP/(TP+FN)$. For the majority class, its precision=$TN/(TN+FN)$, and TNrate=$TN/(TN+FP)$. $F-measure=\frac{((1+\beta^2)*recall+precision)}{(\beta^2*TPrate+precision)}$, where $\beta$ is set to 1 in this paper. In Tables 2 to 7, F-maj and F-min are F-measure for each class. 
\begin{table}[]
\caption{Characteristics of the benchmark datasets.}\label{tab3}
\begin{tabular}{llllll}\hline
\textbf{Dataset}&	\textbf{\#instances}&	\textbf{\#positive instances}&	\textbf{\#attributes}&	\textbf{Positive class label}&	\textbf{IR}   \\\hline
\textbf{Breast\_w}&	699&	241&	9&	Malignant&	1.90\\
\textbf{Diabetes}&	768&	268&	8&	Tested\_positive&	1.86\\
\textbf{Glass}&	214&	17&	9&	3&	11.58\\
\textbf{Ionosphere}&	351&	126&	34&	B&	1.78\\ 
\textbf{Musk}&	476&	208&	168&	Non-Musk&	1.77\\
\textbf{Satimage}&	6430&	625&	36&	2&	9.28\\
\textbf{Segmentation}&	1500&	205&	19&	brickface&	6.31\\
\textbf{Sonar}&	208&	97&	60&	Rock&	1.14\\
\textbf{Vehicle}&	846&	199&	18&	Van&	3.25\\
\textbf{COVID-19 blood test}&	603&	83&	17&	Positive&	6.26\\\hline
\end{tabular}
\end{table}

\begin{figure*}[ht]
\centerline{\includegraphics[width=.9\textwidth,clip=]{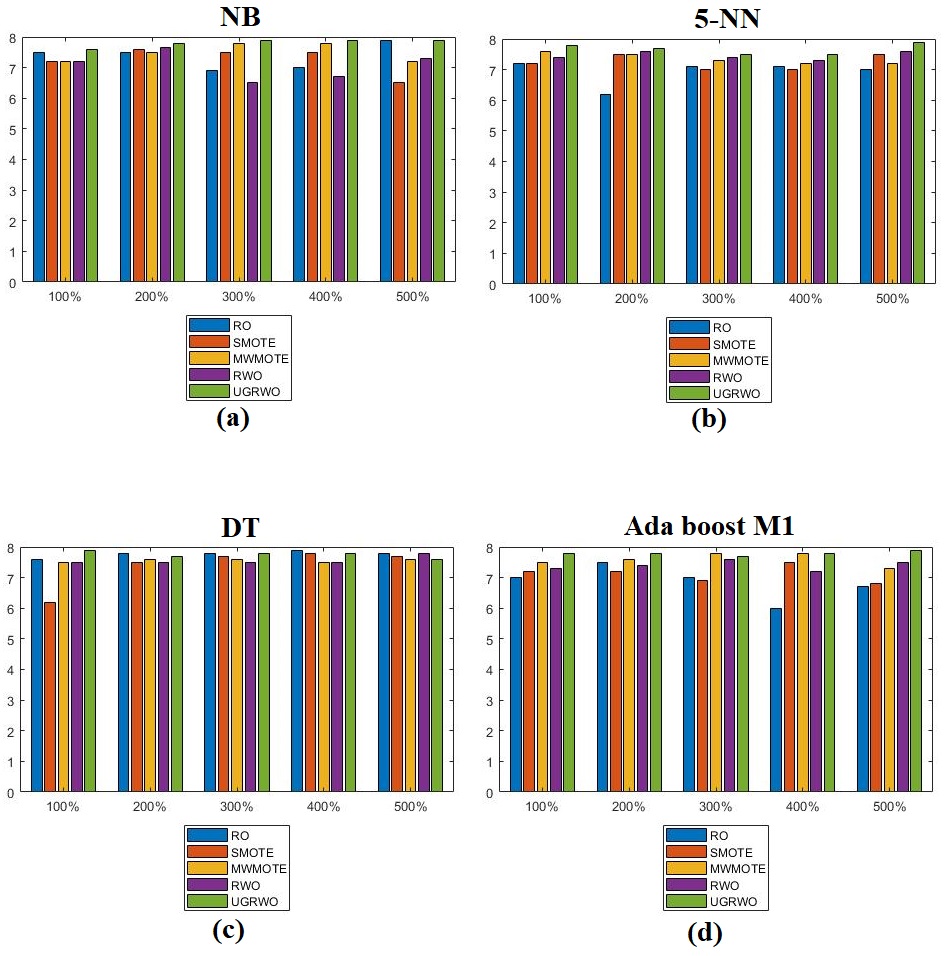}}
\caption{AUCRatio on the ten continuous attribute datasets under five data over-sampling rates and on the original datasets.}\label{fig3}
\end{figure*}

\subsection{Test using UCI and COVID-19 datasets}\label{ssec3_2}
This section will study the following classification approaches; Naïve Bayes, 5-Nearest Neighbor, Decision Tree, GRWO and AdaBoostM1. These five different classifiers should be trained and tested for each dataset, and six different sampling methods should be performed. The GRWO-Sampling is the same as the RWO-Sampling method, except that the RWO-Sampling should be done only on the samples of the minority class, which is selected from the local graph. Tables 5-9 in the Appendix present the implementation of different sampling methods with different over-sampling rates on various datasets.

Table \ref{tab5} shows that when 5-NN is implemented on all datasets, UGRWO-Sampling has the best performance in most evaluation metrics. However, Naïve Bayes and Decision Tree classifiers, which also belong to this method, do not perform well on Vehicle and Glass datasets. With this over-sampling rate, most of the classifiers with UGRWO-Sampling will perform well on the datasets with high Imbalanced Ratios such as Glass and Satimage, but better on the datasets with low Imbalanced Ratios such as Breast\_w and Diabetes because, with increasing over-sampling rate, the balance is established between the two classes. Also, this method has an acceptable performance on the big dataset, e.g., Satimage. In the Satimage data set, there are more outlier samples, which despite the removal mechanism and not considering the outlier samples, a very good improvement has been made after implementing the new UGRWO method on this data set. When implementing the AdaBoostM1 classifier on Musk and Segmentation, RWO-Sampling and RO-sampling have the best performance, and UGRWO-Sampling outperforms best on the other datasets. The RWO method works better on high-dimensional data sets such as Musk than the UGRWO method.

At the end of Table \ref{tab5}, the results of performing all methods with all types of classifiers on the COVID-19 dataset are shown. This dataset has a high imbalance ratio compared to other datasets and is a high-dimensional data set. In this section of the table, the superiority of the UGRWO-sampling method over other methods is obvious. UGRWO-Sampling has the best performance in most evaluation metrics with an over-sampling rate of 100\%.

From the results shown in Table \ref{tab6}, it can be concluded that UGRWO-Sampling performs best on most datasets. However, the proposed sampling method does not perform best on the Glass dataset in all classifiers. In general, similar to the results of Table \ref{tab3}, UGRWO-Sampling on the small dataset with a high Imbalanced Ratio does not perform well. However, it has an extraordinary efficiency on big datasets with a high Imbalanced Ratio because there are more samples and information for sampling in the majority and minority classes. AdaboostM1 classifier with UGRWO-Sampling has an acceptable performance on most datasets except Segmentation and Glass datasets because in these datasets, the number of samples in two-class is low, and the Imbalance Ratio is high. Also, ensemble classifiers such as Adaboost do not perform well and divide the data between different classifiers.

This table shows the superiority of the UGRWO-sampling method in the Naive Bayes and 5-Nearest Neighbors, and in the other classifiers, the MWSMOTE and ROS method are better than the UGRWO in the over-sampling rate 200\%. So far, it can be clearly said that up to an acceptable percentage of over-sampling (100\% and 200\%), the UGRWO-sampling method has a clear advantage in the NB and 5-NN classifiers.

Table \ref{tab7} indicates that RO-Sampling, Tomek+SMOTE, and SMOTE sampling methods perform best in most evaluation metrics when all classifiers are implemented on Glass datasets. The results on the Vehicle dataset show that only if 5-NN and AdaBoostM1 are conducted UGRWO-Sampling will they have the best performance in most evaluation metrics. The results on the Sonar dataset show that the 5-NN classifier has the best performance with RWO-Sampling, but the rest of the classifiers have better efficiency with UGRWO-Sampling. Similar to the results of Tables 5 and 6, UGRWO-Sampling performs best on the Breast\_w, Diabetes, and Satimage data sets. AdaBoostM1 classifier with RO-Sampling and RWO-Sampling had better performance on Musk, Segmentation, and Glass datasets. In general, by increasing the oversampling rate to $300\%$, it is shown that in most classifiers, the proposed method does not work very well on the Glass dataset with a high Imbalance Ratio and a low number of samples because many samples are considered outliers and are excluded from the sampling methods.

According to the end of Table  \ref{tab7}, it can be concluded that the UGRWO method has performed well in all methods on the COVID-19 dataset except the implementation with the decision tree classifier, and in the DT classifier, the MWSOMTE method has performed better than the other methods.

In Table \ref{tab8}, Naive Bayes classifier with RWO-Sampling has an acceptable performance on Diabetes, Segmentation, and Vehicle datasets, Decision Tree classifier performs well with RO-Sampling, and MWMOTE has the best efficiency on Diabetes, Ionosphere and Segmentation data sets. All the classifiers except 5-NN with UGRWO-Sampling had better performance on the Sonar dataset because the Sonar data set have a low Imbalance Ratio and several samples, so new samples in UGRWO do not have a relation with others very well. UGRWO-Sampling did not have acceptable performance on the Glass dataset though, RO-Sampling performed well in the Decision Tree classifier on most of the datasets. When implementing the AdaBoostM1 classifier on Musk, Segmentation, and Glass, RO-Sampling, Tomek+SMOTE, and SMOTE outperform best UGRWO-Sampling has the best performance on other datasets. The UGRWO method on data sets with a low number of samples and an Imbalance Ratio in a high over-sampling rate is not very good because many samples in the majority class are excluded from the method, and also, information from data is not enough for sampling.

In this table's result of the COVID-19 dataset, the UGRWO-sampling method has a good primacy in 5-NN, ADA-Boost classifiers. However, in other classifiers, other methods have an acceptable value in Tp rate, G-mean evaluation method, so other methods in this size of over-sampling have a good advantage in imbalance because these criteria Show the superiority of the method at high imbalance rates.

From the results presented in Table \ref{tab9}, UGRWO-Sampling performs best on the Breast\_w and Satimage datasets. This suggests that the UGRWO method improves accuracy in big datasets by increasing the oversampling rate because efficiency increases by removing outliers and balancing between the two classes. RO-Sampling is useful when all classifier is implemented on Glass datasets. 5-NN classifier with RWO-Sampling best on the Sonar dataset and other classifiers with UGRWO-Sampling. When implementing a Decision Tree classifier on Diabetes, Ionosphere, and Vehicle, RO-Sampling outperforms the other three approaches, and UGRWO-Sampling outperforms best in the other classifiers. When AdaBoostM1 is implemented, RO-Sampling, Tomek+SMOT, and SMOTE have the best performance on Musk, Segmentation, and Glass datasets.

According to the end of this table, it can be concluded that the UGRWO-sampling method is very suitable in the 5-NN and ADA-Boost classifiers on the COVID-19 dataset, and in other classifiers, other methods are more efficient on this dataset.

Generally, as shown in Tables 5-9, the results on the large scale and highly imbalanced datasets, Satimage, show that most of the time, UGRWO-Sampling performs best in most metrics. Sonar and Musk are the highest-dimensional sample datasets, and RWO-Sampling and MWMOTE always have the best efficiency in 300, 400, and 500\% of over-sampling rates, and UGRWO-Sampling performs well in all metrics in 100 and 200\% of over-sampling rates. This shows that the proposed UGRWO method works well on small datasets at a high sampling rate to some extent. However, efficiency will decrease with an increased sampling rate on such datasets due to insufficient information to increase and absorb outliers. Therefore, the RWO method will work better. Glass with the lowest-dimensional sample, high Imbalanced Ratio dataset, RO-Sampling, Tomek+SMOTE, and SMOTE always perform well under all over-sampling rates. The results of Breast\_w, Diabetes, and Ionosphere datasets show that our approach performs well in most metrics.

When the 5-NN and NB  classifiers are implemented on the Segmentation dataset, UGRWO-Sampling has the best efficiency in all metrics except in 400\% of the sampling rate for NB. Furthermore, when Decision Tree and AdaBoostM1 classifiers are implemented, RO-Sampling, Tomek+SMOTE, and SMOTE have the best performance. Also, when Naive Bayes is implemented on the Vehicle dataset, RWO-Sampling always has the best efficiency. However, when the over-sampling rate is 300\%, AdaBoostM1 with UGRWO-Sampling does not have acceptable performance in all metrics.

Generally, by increasing the oversampling rate, the efficiency of the proposed method decreases on different datasets. However, this method performs well on large-scale datasets. Therefore, it can be concluded that the proposed method can work well in almost all cases on low Imbalanced Ratio datasets. It should be noted that the method has also worked well in high Imbalanced Ratio datasets such as Satimage and Segmentation, but on Glass dataset, SMOTE, Tomek+SMOTE, and RO-Sampling methods have better performance in all classifiers. The SMOTE and Tomek+SMOTE methods generally work better on low-dimensional datasets with a smaller volume than the UGRWO method because the UGRWO method uses the under-sampling method to remove outliers, resulting in several data samples. Therefore, the low density will decrease, and few samples will remain, but the UGRWO method will work better on data sets of medium to high dimensional and sizes.

Paying special attention to the bottom of the tables, the superiority of the UGRWO-sampling method in 5-NN and ADA-Boost classifiers can be seen in all over-sampling rates. In other classifiers, it can be seen that the new method with percentages (100 and 200) over-sampling still works well on this dataset and was able to diagnose people with COVID-19 disease correctly. However, with the increasing percentage of over-sampling rates, the rest of the methods have surpassed the new method due to increased outlier and lack of sufficient data from the dataset for sampling.

In order to compare the performance conveniently, we counted the number of wins in all cases for each over-sampling approach and provided the results in Table 4. The results show that UGRWO-Sampling outperforms the other five approaches in all six metrics when conducting Naive Bayes, 5-NN, and Decision Tree, and it loses to GROW-Sampling only once F-major. The results also reveal that GRWO-Sampling and MWMOTE are the most time-consuming methods. This indicates that in some low-scale datasets, despite the under-sampling, much data is lost, and the training is not done well, so despite the under-sampling mechanism in the UGRWO method, many majority samples will also disappear, and many data will be discarded. In contrast, an ensemble classifier that gives part of the data to each individual classifier exacerbates the problem. Based on the results presented in Table 4, we can conclude that UGRWO-Sampling performs potentially well in imbalanced data sets no matter what classification algorithms are conducted.

\begin{table}[h]
\caption{Five evaluation metric win summary on ten continuous attribute datasets when implementing four baseline classifiers.}\label{tab4}
\begin{tabular}{llllllll}\hline
\textbf{Alg}&	\textbf{OS}&	\textbf{F-min}&	\textbf{F-maj}&	\textbf{acc}&	\textbf{G-mean}&\textbf{TPrate}&\textbf{AUC}   \\\hline
\textbf{NB}&	UGRWO&	\textbf{39}&	\textbf{29}&	\textbf{32}&	\textbf{29}&	\textbf{27}&	\textbf{29}\\
&	GRWO&	1&	11&	0&	1&	6&	1\\
&	RWO&	7&	7&	7&	9&	3&	6\\
&	MWMOTE&	3&	1&	1&	3&	4&	3\\
&	SMOTE&	1&	2&	1&	0&	2&	3\\
&	Tomek+SMOTE&	1&	1&	1&	0&	3&	3\\
&	ROS&	4&	1&	1&	5&	4&	1\\
\textbf{5-NN}&	UGRWO&	\textbf{35}&	\textbf{18}&	\textbf{33}&	\textbf{29}&	\textbf{37}&	\textbf{34}\\
&	GRWO&	0&	10&	0&	2&	0&	0\\
&	RWO&	6&	4&	7&	8&	6&	4\\
&	MWMOTE&	0&	1&	0&	0&	0&	2\\
&	SMOTE&	1&	8&	2&	3&	6&	2\\
&	Tomek+SMOTE&	1&	4&	2&	4&	3&	2\\
&	ROS&	8&	4&	7&	4&	14&	1\\
\textbf{DT}&	UGRWO&	\textbf{26}&	\textbf{14}&	\textbf{23}&	\textbf{15}&	\textbf{15}&	\textbf{24}\\
&	GRWO&	0&	2&	2&	0&	0&	5\\
&	RWO&	5&	4&	3&	5&	4&	4\\
&	MWMOTE&	5&	8&	5&	5&	6&	7\\
&	SMOTE&	0&	3&	2&	3&	2&	4\\
&	Tomek+SMOTE&	0&  3&	2&	3&	3&	3\\
&	ROS&	9&	9&	10&	8&	11&	1\\
\textbf{Ada}& UGRWO&	\textbf{37}&	16&	\textbf{33}	& \textbf{32}&	\textbf{31}&	\textbf{28}\\
&	GRWO&	1&	\textbf{19}&	1&	7&	0&	2\\
&	RWO&	2&	4&	2&	6&	1&	6\\
&	MWMOTE&	1&	3&	2&	3&	2&	4\\
&	SMOTE&	2&	2&	3&	1&	3&	6\\
&	Tomek+SMOTE&	2&	1&	1&	2&	1&	5\\
&	ROS&	7&	6&	7&	6&	7&	2\\\hline
\end{tabular}
\end{table}

AUC (Area Under the Curve) is also an important metric to evaluate the performance of classifiers. According to the above results, the proposed method had an acceptable performance in most datasets, but we cannot claim which classifier has the best performance. The summary in Table \ref{tab4} shows that UGRWO-Sampling outperforms the others statistically in terms of AUC.

For a better conclusion, we use another measure called AUCRatio. At first, we computed the relative performance of a given method $M$ on a dataset $i$ as the ratio between its AUC and the highest among all the compared methods,

\begin{equation*}
    AUCRatio_{i}(M)=\frac{AUC(M)}{max_{j} AUC(j)}.
\end{equation*}
Where AUC(j) is the AUC for method j on the data set i, the larger the value of AUCRatioi(M), the better the performance of M in data set i \cite{Kubat_1997}. Figure \ref{fig3} depicts the distribution of the relative performance of the six methods and all data sets. 

According to Figure \ref{fig3}, the proposed method has proved to have acceptable performance in comparison to other methods. AUCRatio shows that UGRWO-Sampling has outperformed the RWO-Sampling. The proposed method deals with a continuous attribute value to create synthetic samples and will enhance the efficiency of the RWO method. As shown in Figure \ref{fig3}, it can be concluded that in all classifications at the lower over-sampling rate, the proposed method would have better efficiency. In all cases, when Naive Bayes and AdaBoostM1 classifiers are in operation, the proposed method boosts their performance. When the 5-NN classifier is at the operation level, UGRWO-Sampling and RWO-Sampling outperform the other approaches, respectively. Decision Tree with RO-Sampling also has a great performance, but UGRWO-Sampling remains better than RO-Sampling.

\section{Conclusion and future works}\label{sec4}
We have evaluated four classifiers, including the Naive Bayes classifier, K-Nearest Neighbor, Decision Tree, and AdaBoostM1 on imbalanced data sets. Over-Sampling often affects the performance of k nearest neighbors (KNN) since prior to over-sampling, the number of the minority class instances in the fixed volume may be increased.

A Naive Bayes classifier obtains the posterior probability for a test sample, and over-sampling increases the posterior probability of the minority class data; thus, over-sampling influences the performance. Over-sampling often affects the performance of the Decision Tree in response to the modification of data distribution since it influences the measure called information gain used for choosing the best attribute in the Decision Tree and consequently leads to the modification of the constructed Decision Tree. Modifying the structure of the Decision Tree influences pruning and over-fitting avoidance and consequently influences the performance of the Decision Tree. Also, Ada, Boost, and M1 change the underlying data distribution and classify the re-weighted data space iteratively.

RO-Sampling is simple, but it increases the possibility of over-fitting. SMOTE uses linear interpolation for sampling generation, and new samples fall on the line segment connected by two neighbors. It does not expand the space occupied by the minority class data, changes the original data distribution of the minority class data, and does not change the original data distribution of the minority class. When generating synthetic samples, RWO-Sampling tries to keep the minority class data distribution unchanged while expanding the space occupied by the minority class data. Thus it has high generalization capability and performs well on imbalanced data classification. RWO-Sampling does not generate synthetic samples around the mean point of the real minority class data through the random walk model since it also tends to increase the likelihood of over-fitting, but the proposed method has attempted to find solutions for these problems. Therefore, using the KNN method, the UGRWO-Sampling tends to create synthetic samples around the mean since by eliminating the impact of noise or outlier samples, new samples will be generated around the mean. Also, compared to RWO-Sampling, this method is less likely to increase the probability of over-fitting, and it has high generalizability. However, due to the use of the KNN method for the proposed method and being aware that KNN would fail on a large scale, the proposed method would have a quite weak performance on a large scale.

RO-Sampling is less consuming in comparison with other approaches in generating new samples. Tomek+SMOTE and SMOTE need to calculate the K nearest neighbors for a chosen sample before generating new samples, and RWO-Sampling needs to calculate the mean and standard deviation for all attributes. K nearest neighbors are time-consuming compared to the mean and standard deviation calculation, so RWO-Sampling acts faster than SMOTE. On the other hand, the tomek+SMOTE method needs to perform SMOTE and Tomek together, so this method runs slower. In the new method, the K nearest neighbors, mean and standard deviation for all attributes must be calculated. Therefore, compared to SMOTE and RWO-Sampling methods, UGRWO-Sampling is more time-consuming.

Paying special attention to the COVID-19 dataset, it can be concluded that the UGRWO-sampling method works very well on a variety of classifiers at over-sampling rates of 100 to 300, but with increasing sampling rates, other methods have shown better results, somewhat later in the evaluation metric. Therefore, it can be concluded that the UGRWO-sampling method in diagnosing COVID-19 disease by choosing the appropriate over-sampling rate has performed much better than the compared methods and has had a very good diagnosis of this type of disease.

A new combined method is proposed to reduce the disadvantages of over-sampling and under-sampling methods by which data sets perform well in most cases (e.g., in continuous data sets), though not performing well on discrete data sets, and this issue would benefit from further research.

 \newpage
 \section{Appendix}
\begin{longtable}{llllllll}
\caption{Averaged results and standard deviations on ten continuous attribute data sets (over-sampling rate equals $100\%$).}
\label{tab5}
\endfirsthead
\endhead
\hline
 ds	& Alg	& OS	& f-min	& f-maj	& O-acc	& G-mean	& TP rate  \\\hline
\textbf{Breast\_w}&	NB&	UGRWO&	\textbf{97.01$\pm$0.01}&	\textbf{96.98$\pm$0.03}&	\textbf{96.53$\pm$0.05}&	\textbf{97.50$\pm$0.56}&	\textbf{97.54$\pm$0.45} \\
& &GRWO&	96.62$\pm$0.01&	96.78$\pm$0.01&	96.23$\pm$2.14&	97.30$\pm$3.16&	96.25$\pm$2.37\\
& &RWO&	96.38$\pm$0.01&	96.15$\pm$0.01&	96.26$\pm$1.52&	96.23$\pm$1.52&	97.28$\pm$2.70\\
& &MWMOTE&	95.24$\pm$0.02&	96.52$\pm$0.01&	96.42$\pm$0.05&	96.05$\pm$0.32&	97.26$\pm$0.21\\
& &SMOTE&	94.18$\pm$0.00&	96.77$\pm$0.00&	95.85$\pm$0.81&	96.10$\pm$0.70&	97.10$\pm$3.42\\
& &Tomek+SMOTE&	95.08$\pm$0.00&	95.59$\pm$0.00&	96.00$\pm$0.63&	95.95$\pm$1.64&	96.04$\pm$2.62\\
& &ROS&	96.52$\pm$0.02&	96.24$\pm$0.02&	96.39$\pm$2.17&	96.35$\pm$2.20&	97.52$\pm$1.87\\
\end{longtable}

\begin{longtable}{llllllll}
\caption{Averaged results and standard deviations on ten continuous attribute datasets (over-sampling rate equals $200\%$).}
\label{tab6}
\endfirsthead
\endhead
\hline
 ds	& Alg	& OS	& f-min	& f-maj	& O-acc	& G-mean	& TP rate  \\\hline
 Breast\_w&	NB&	UGRWO&	\textbf{98.23$\pm$0.03}&	\textbf{96.99$\pm$0.03}&	\textbf{96.98$\pm$2.03}&	\textbf{96.52$\pm$2.04}&	\textbf{98.01$\pm$1.34}\\
&&		GRWO&	95.52$\pm$0.02&	96.27$\pm$0.02&	96.08$\pm$2.23&	96.33$\pm$2.11&	97.51$\pm$2.40\\
&&		RWO&	97.36$\pm$0.01&	95.83$\pm$0.01&	96.77$\pm$1.48&	96.47$\pm$1.64&	97.77$\pm$1.49\\
&&		MWMOTE&	97.25$\pm$0.01&	95.26$\pm$0.02&	96.52$\pm$0.25&	95.25$\pm$6.25&	97.48$\pm$2.54\\
&&		SMOTE&	96.17$\pm$0.01&	95.96$\pm$0.01&	96.06$\pm$1.50&	96.03$\pm$1.47&	96.89$\pm$2.81\\
&&	Tomek+SMOTE &	97.16$\pm$1.16&	96.68$\pm$0.84&	96.56$\pm$1.59&96.65$\pm$0.59&	96.56$\pm$0.16\\
&&		ROS&	97.45$\pm$0.01&	95.92$\pm$0.02&	96.86$\pm$1.26&	96.54$\pm$1.49&	97.92$\pm$0.97\\
\end{longtable}
 \begin{longtable}{llllllll}
\caption{Averaged results and standard deviations on ten continuous attribute datasets (over-sampling rate equals $300\%$).}
\label{tab7}
\endfirsthead
\endhead
\hline
 ds	& Alg	& OS	& f-min	& f-maj	& O-acc	& G-mean	& TP rate  \\\hline
Breast\_w&	NB&	UGRWO&	\textbf{99.23$\pm$0.02}&	\textbf{96.25$\pm$0.01}&	\textbf{97.85$\pm$0.02}&	\textbf{96.84$\pm$2.05}&	\textbf{98.54$\pm$2.35}\\
&&		GRWO&	95.94$\pm$0.03&	96.86$\pm$0.02&	96.31$\pm$2.38&	96.57$\pm$2.12&	98.00$\pm$1.72\\
&&		RWO&	97.90$\pm$0.01&	95.61$\pm$0.02&	97.16$\pm$1.46&	96.63$\pm$1.73&	98.11$\pm$1.28\\
&&		MWMOTE&	98.25$\pm$0.02&	95.26$\pm$0.02&	96.25$\pm$0.03&	95.46$\pm$1.02&	97.15$\pm$4.02\\
&&		SMOTE&	97.39$\pm$0.02&	95.83$\pm$0.03&	96.79$\pm$2.59&	96.48$\pm$2.78&	97.79$\pm$2.26\\
&&	Tomek+SMOTE &	97.28$\pm$0.95&	96.05$\pm$0.24&	96.00$\pm$1.05&96.26$\pm$0.84&	97.05$\pm$0.24\\
&&		ROS&	98.03$\pm$0.00&	95.83$\pm$0.01&	97.32$\pm$0.64&	96.74$\pm$0.87&	98.34$\pm$1.21\\
\end{longtable}

\begin{longtable}{llllllll}
\caption{Averaged results and standard deviations on ten continuous attribute datasets (over-sampling rate equals $400\%$).}
\label{tab8}
\endfirsthead
\endhead
\hline
 ds	& Alg	& OS	& f-min	& f-maj	& O-acc	& G-mean	& TP rate  \\\hline
 Breast\_w&	NB&	UGRWO&	\textbf{98.87$\pm$0.01}&	\textbf{97.02$\pm$0.03}&	\textbf{97.81$\pm$2.14}&	\textbf{96.98$\pm$2.06}&	\textbf{98.45$\pm$2.04}\\
&&		GRWO&	95.96$\pm$0.02&	96.88$\pm$0.01&	96.41$\pm$1.45&	96.63$\pm$1.35&	98.12$\pm$2.18\\
&&		RWO&	98.28$\pm$0.01&	95.52$\pm$0.02&	97.51$\pm$1.52&	96.78$\pm$2.06&	98.40$\pm$1.28\\
&&		MWMOTE&	97.58$\pm$0.02&	92.03$\pm$0.02&	96.25$\pm$0.01&	95.15$\pm$0.12&	98.02$\pm$0.02\\
&&		SMOTE&	97.55$\pm$0.01&	94.90$\pm$0.02&	96.69$\pm$1.56&	96.29$\pm$1.73&	97.40$\pm$1.64\\
&&	Tomek+SMOTE &	97.00$\pm$0.25&	94.00$\pm$0.26&	96.35$\pm$1.16&96.16$\pm$0.03&	97.05$\pm$0.16\\
&&		ROS&	98.00$\pm$0.00&	94.77$\pm$0.02&	97.11$\pm$1.19&	96.49$\pm$1.91&	97.84$\pm$1.12\\
\end{longtable}
\begin{longtable}{llllllll}
\caption{Averaged results and standard deviations on ten continuous attribute datasets (over-sampling rate equals $500\%$).}
\label{tab9}
\endfirsthead
\endhead
\hline
 ds	& Alg& OS	& f-min	& f-maj	& O-acc	& G-mean	& TP rate  \\\hline
Breast\_w&	NB&	UGRWO&	\textbf{98.85$\pm$0.01}&	\textbf{97.04$\pm$0.02}&	\textbf{97.85$\pm$2.14}&	\textbf{96.98$\pm$1.48}&	\textbf{98.55$\pm$0.08}\\
&&		GRWO&	96.36$\pm$0.01&	96.99$\pm$0.02&	96.62$\pm$2.50&	96.83$\pm$2.44&	98.52$\pm$2.85\\
&&		RWO&	98.32$\pm$0.00&	94.79$\pm$0.02&	97.46$\pm$1.33&	96.67$\pm$2.05&	98.18$\pm$1.10\\
&&		MWMOTE&	98.37$\pm$0.01&	96.91$\pm$0.01&	96.16$\pm$1.25&	95.96$\pm$1.47&	92.47$\pm$0.83\\
&&		SMOTE&	97.74$\pm$0.00&	94.18$\pm$0.02&	96.75$\pm$1.39&	96.25$\pm$1.77&	97.34$\pm$1.50\\
&&	Tomek+SMOTE &	97.36$\pm$0.16&	94.00$\pm$0.14&	96.16$\pm$1.16&96.13$\pm$0.16&	97.05$\pm$0.17\\
&&		ROS&	98.33$\pm$0.00&	94.80$\pm$0.02&	97.48$\pm$1.07&	96.68$\pm$1.56&	98.20$\pm$1.35\\
\end{longtable}

\bibliography{library}
\end{document}